\documentclass[conference]{IEEEtran}
\usepackage{cite}
\usepackage{graphicx}
\usepackage{url}
\usepackage{amsmath}
\usepackage[caption = false, font=footnotesize]{subfig}
\usepackage{float}
\usepackage{bm}
\usepackage{booktabs} 
\usepackage{multirow}
\usepackage{pifont}
\usepackage{array}
\usepackage{color}
\usepackage{tabularx}
\usepackage{longtable}
\usepackage{tabu}
\newcolumntype{L}[1]{>{\raggedright\let\newline\\\arraybackslash\hspace{0pt}}m{#1}}
\newcolumntype{C}[1]{>{\centering\let\newline\\\arraybackslash\hspace{0pt}}m{#1}}
\newcolumntype{R}[1]{>{\raggedleft\let\newline\\\arraybackslash\hspace{0pt}}m{#1}}
\newcommand{\etal}{\textit{et al}.}
\newcommand{\ie}{\textit{i}.\textit{e}.}
\newcommand{\eg}{\textit{e}.\textit{g}.}

\usepackage{caption2}

\usepackage{graphicx,color, multirow, amsmath}

\RequirePackage{latexsym,amsmath,amssymb}

\def\BibTeX{{\rm B\kern-.05em{\sc i\kern-.025em b}\kern-.08em
    T\kern-.1667em\lower.7ex\hbox{E}\kern-.125emX}}

\begin{document}

\title{Perceptually Optimized Deep High-Dynamic-Range Image Tone Mapping
}

\author{\textit{Chenyang Le$^1$, Jiebin Yan$^1$, Yuming Fang$^1$, Kede Ma$^2$} \\
 $^1$School of Information Management, Jiangxi University of Finance and Economics, Nanchang, China \\
 $^2$ Department of Computer Science, City University of Hong Kong, Kowloon, Hong Kong \\}

\maketitle

\begin{abstract}
We describe a deep high-dynamic-range (HDR) image tone mapping operator that is  computationally efficient and perceptually optimized. We first decompose an HDR image into a normalized {Laplacian} pyramid, and use two deep neural networks (DNNs) to estimate the {Laplacian} pyramid of the desired tone-mapped image from the normalized representation. We then end-to-end optimize the entire method over a database of HDR images by minimizing the normalized {Laplacian} pyramid distance (NLPD), a recently proposed perceptual metric. Qualitative and quantitative experiments demonstrate that our method produces images with better visual quality, and runs the fastest among existing local tone mapping algorithms.
\end{abstract}


\section{Introduction}
\begin{figure}[t]
        \centering
        \subfloat[Linear scaling]{
            \includegraphics[width=0.23\textwidth]{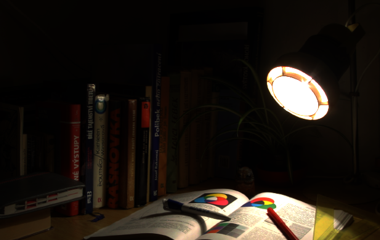}
        }
        \subfloat[Drago03~\cite{drago2003adaptive}]{
            \includegraphics[width=0.23\textwidth]{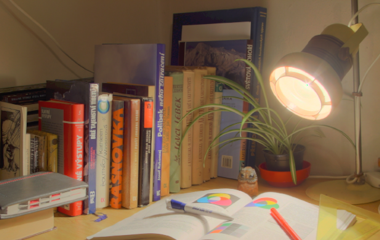}
        }\\
        \vspace{-0.3cm}
        \subfloat[WLS~\cite{farbman2008edge}]{
            \includegraphics[width=0.23\textwidth]{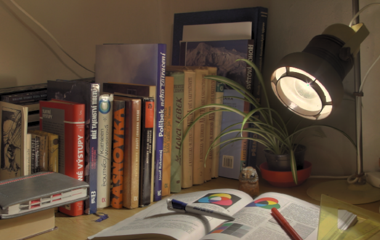}
        }
        \subfloat[Ours]{
            \includegraphics[width=0.23\textwidth]{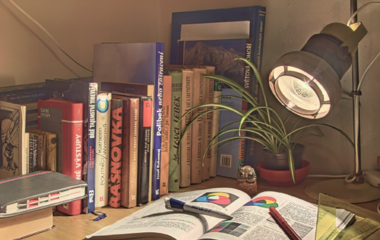}
        }
        \caption{Tone mapping results of the ``Lamp'' image  courtesy of Martin \v{C}ad\'{i}k.}
		    \label{fig:lamp}
    \end{figure}

 Existing monitors, projectors, and print-outs have a significantly limited dynamic range, which is inadequate to reproduce the full spectrum of luminance values presented in natural scenes and captured by current sensors~\cite{reinhard2010high}. When rendering high-dynamic-range (HDR) images on low-dynamic-range (LDR) display devices, tone mapping operators (TMOs) are necessary for dynamic range compression, preserving salient visual features of the original scenes. A na\"{i}ve TMO is to \textit{linearly} rescale the luminance values to  the displayable range. However, this method is sensitive to the maximum luminance of a scene, and often creates a dark appearance (see Fig.~\ref{fig:lamp} (a)). In the past decade, a large number of \textit{non-linear} TMOs~\cite{reinhard2002photographic,durand2002fast,li2005compressing,meylan2006high,farbman2008edge} have been proposed, aiming for faithful tone reproduction and detail preservation. These can be broadly classified into two categories: global and local operators. Global TMOs~\cite{ward1994radiance,larson1997visibility,tumblin1993tone,drago2003adaptive,reinhard2005dynamic,kim2008consistent} are a set of parametric functions, including homography, gamma mapping, logarithmic function~\cite{drago2003adaptive}, and sigmoid non-linearity~\cite{reinhard2005dynamic}. Global methods preserve global contrast well, but may lose local details.
    Recent studies mainly focus on local TMOs~\cite{durand2002fast,farbman2008edge,paris2011local,BRUCE201412,shibata2016gradient,liang2018hybrid}. A common theme is to  decompose an HDR image into a base layer and a detail layer. Tone mapping is applied to the base layer, while detail enhancement  is done in the detail layer. Along this path, many methods~\cite{farbman2008edge,shibata2016gradient,liang2018hybrid} have been proposed,  differing mainly in how to perform the two-layer image decomposition in a ``more effective'' way. Local methods usually produce images with satisfactory local contrast with improved visual quality. However, this often comes at the cost of increasing computational complexity~\cite{paris2011local}. In addition, global contrast may be reduced, and localized artifacts such as halo-like glowing may emerge in the tone-mapped images. 
	
	\begin{figure*}[t]
    \centering
    \includegraphics[scale=0.65]{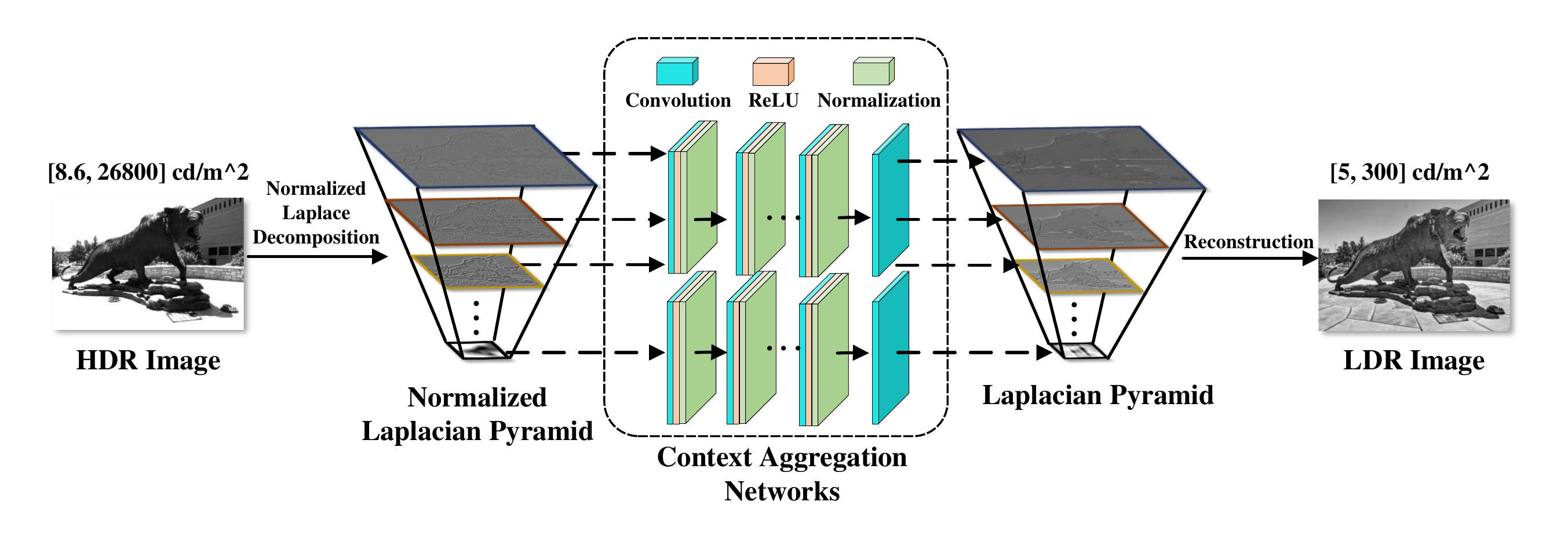}
    \vspace{-.2cm}
    \caption{Schematic diagram of the proposed TMO.}
    \label{fig:framework}
\end{figure*} 

	Perceptual optimization of HDR image tone mapping was investigated by Yeganeh and Wang~\cite{yeganeh2013high}. They searched over the space of all feasible tone-mapped images for the closest one with respect to the original scene, measured by a structural fidelity index~\cite{yeganeh2012objective}. This method was later improved in~\cite{ma2015high} by incorporating a statistical naturalness measure. Laparra~\etal~\cite{laparra2017perceptually} formulated HDR image tone mapping as a more general image rendering problem, taking into account various display constraints. Nevertheless, the above methods require gradient-based iterative optimizers operating in the high-dimensional spaces, which are computationally expensive, preventing their wide adoption in real-world applications.

This paper aims to develop a TMO for rendering HDR images with two desired design principles. First, it should be \textit{computationally efficient}. We first decompose the input HDR image into a normalized Laplacian pyramid \cite{burt1983laplacian,laparra2017perceptually}. Instead of iteratively optimizing over the space of all feasible tone-mapped images, we train two feed-forward deep neural networks (DNNs): one accepts all bandpass channels and the highpass channel, and the other processes the lowpass channel of the normalized representation. Together, they predict the Laplacian pyramid of the desired tone-mapped image.  The two networks are designed to be highly light-weight, enabling our method to run the fastest among existing \textit{local} TMOs. Second, it should be \textit{perceptually optimized}. Unlike most TMOs, we end-to-end train our networks by optimizing a recently proposed perceptual metric - the  normalized Laplacian pyramid distance (NLPD)~\cite{laparra2017perceptually} between the predicted LDR images and the corresponding HDR scenes.  Experiments on a test set of HDR images show that the optimized method performs consistently better than existing TMOs both qualitatively and quantitatively (measured by an independent perceptual metric - TMQI \cite{yeganeh2012objective}).
    
\section{The NLPD Metric}
\label{sec:related_work}
In this section, we provide a brief review of the NLPD metric~\cite{laparra2017perceptually}, which will be adopted as the learning objective of the proposed DNN-based TMO.

 NLPD is motivated by the physiology of the early visual system. Given a calibrated HDR image $S$,  the luminance values (in the unit of candela  per  square  meter, cd/m$^2$) are firstly preprocessed by an exponential function, approximating the transformation of light to the response of retinal photoreceptors~\cite{laparra2017perceptually}
\begin{align}
    x^{(1)} = S^\gamma.\label{eq:gamma}
\end{align}
Luminance subtraction and contrast normalization are then applied recursively to partition $x^{(1)}$ into frequency subbands,  mimicking the center-surround receptive fields found in the retina and the lateral geniculate nucleus~\cite{laparra2017perceptually}
\begin{align}
      &x^{(i+1)} = DLx^{(i)}, \quad i\in\{1,\ldots,m-1\}, \label{eq:n1}\\
      &z^{(i)} = x^{(i)} -LUx^{(i+1)},\\
      &z^{(m)} = x^{(m)},
\end{align}
where $D$ and $U$ represent linear  down-/up-sampling by a factor of two, respectively. The lowpass filter $L$ is inherited from the Laplacian pyramid~\cite{burt1983laplacian}. $m$ represents the number of pyramid levels. We obtain the normalized Laplacian pyramid by dividing each coefficient by a weighted sum of neighbouring coefficients within each subband 
\begin{align}
      y^{(i)} = z^{(i)} \oslash (P\vert z^{(i)}\vert + c),\label{eq:normalization}
\end{align}
where $\oslash$ denotes the Hadamard division and $P$ is a convolution filter optimized to reduce statistical dependencies~\cite{laparra2017perceptually}. $c$ is a small positive constant to avoid potential division by zero. Based on the normalized Laplacian pyramid representations
\begin{align}
    f(S)=\{y^{(i)}\}_{i=1}^m \mbox{ and } f(I)=\{\tilde{y}^{(i)}\}_{i=1}^m,
\end{align} 
where $\tilde{y}^{(i)}$ denotes the $i$-th level of the normalized Laplacian pyramid of the tone-mapped image $I$. The final NLPD metric is computed by
\begin{align}
      \ell(S,I) = \left[\frac{1}{m}\sum_{i=1}^{m}\left(\frac{1}{n^{(i)}}\sum_{j=1}^{n^{(i)}}|y^{(i)}_j - \tilde{y}^{(i)}_j|^{\alpha}\right)^{\frac{\beta}{\alpha}}\right]^{\frac{1}{\beta}},
\end{align}
where $n^{(i)}$ is the number of coefficients in the $i$-th channel. The two exponents $\alpha$ and $\beta$ are optimized to match the human perception of image quality using  subject-rated image quality databases. NLPD is continuous and differentiable~\cite{laparra2017perceptually}, which permits gradient-based optimization.

\section{Proposed Method}
\label{sec:proposed_method}

In this section, we describe the proposed TMO. After preprocessing, we decompose the input HDR image into a normalized {Laplacian} pyramid and feed it to two DNNs for Laplacian pyramid estimation, which is further collapsed to obtain the final LDR image. Fig.~\ref{fig:framework} shows the general framework.

\subsection{Preprocessing}
\label{sec:preprocessing}

It is important for TMOs to work with calibrated HDR images, \ie, images with true luminance values for all pixels. Calibration allows TMOs to differentiate between bright and dim scenes. Otherwise, a night HDR image in arbitrary units may be tone-mapped to a day-lit scene with amplified sensor noise. However in practice, many HDR images are acquired without calibration, meaning that the recorded measurements $R$ are linearly proportional to the actual luminances $S$ by an unknown scaling factor. To apply the proposed TMO to an uncalibrated HDR image, we need to make some educated guesses of the minimum and maximum luminance values in the original scene~\cite{laparra2017perceptually}, denoted by $S_{\mathrm{min}}$ and $S_{\mathrm{max}}$, respectively. For example, a typical photographic scene in full sunlight has a luminance about $5\times 10^3$ cd/m$^2$, while a frosted incandescent light bulb is roughly $10^5$ cd/m$^2$. After that, we linearly rescale the measurements to the estimated luminance values
\begin{align}
     & \bar{R} = \frac{R - R_{\mathrm{min}}}{R_{\mathrm{max}}-R_{\mathrm{min}}}\in[0,1], \\
     &S = (S_{\mathrm{max}} -S_{\mathrm{min}})\cdot \bar{R} + S_{\mathrm{min}}.
  \end{align}
 As the final step of preprocessing, we decompose the ``calibrated'' $S$ into the normalized Laplacian pyramid according to Eqs. \eqref{eq:n1} to \eqref{eq:normalization}.
    
    
\begin{table}[t]
	\centering
	\caption{Specification of the two CANs in the proposed TMO. Exclusion of the bias terms makes our method locally scale-invariant, which improves generalization to unseen luminance levels.}
	\vspace{1em}
	\begin{tabular}{l|c c c c c c c}
		\toprule
		Layer & 1 & 2 & 3 & 4 \\
		\midrule
        Convolution & 3 & 3 & 3 & 3  \\
		Dilation & 1 & 2 & 4 & 1 \\
		Width & 32 & 32 & 32 & 1 \\
		Bias & \ding{55} & \ding{55}  & \ding{55} & \ding{55} \\
		Adaptive Normalization & \ding{51} & \ding{51} & \ding{51} & \ding{55} \\
		LReLU Non-linearity & \ding{51} & \ding{51} & \ding{51} & \ding{55}\\
		\bottomrule
	\end{tabular}
	\label{tab:CAN_conf}
\end{table}


\subsection{Network Architecture}
\label{network}

At the core of our method are two DNNs that predict the Laplacian pyramid of the LDR image using the normalized representation of the corresponding  HDR image as input. We choose the context aggregation network (CAN) \cite{yu2015multi} as our default architecture due to its effectiveness in aggregating global context information without spatial resolution reduction.   Table~\ref{tab:CAN_conf} shows the detailed specifications, which are manually optimized to be highly light-weight while balancing the visual quality of the output images. The CAN shared by all bandpass channels and the highpass channel has four convolution layers. Similar in \cite{chen2017fast}, we use adaptive normalization, a combination of the identity mapping and the batch normalization, right after the first three convolutions:
\begin{align}
  \mathrm{AN}(y) = \lambda_1y + \lambda_2 \mathrm{BN}(y),
\end{align}
where $\lambda_1$ and $\lambda_2$ are two learnable parameters.
The weight sharing across bandpass and highpass channels enables accepting a normalized Laplacian pyramid of arbitrary levels. The leaky rectified linear unit (LReLU) is employed as the nonlinear activation function:
\begin{align}
    \mathrm{LReLU}(y) = \max(\lambda_3y,y),
\end{align}
where $\lambda_3 \ge 0$ is a fixed parameter during training. We use another CAN with the same architecture to compress the dynamic range of the lowpass luminance channel. Together the two CANs output the Laplacian pyramid of the desired tone-mapped image, which is  constrained to have a luminance range of $[5, 300]$ cd/m$^2$.

\begin{figure*}[t]
    \subfloat[Drago03~\cite{drago2003adaptive}]{
        \includegraphics[width=0.24\textwidth]{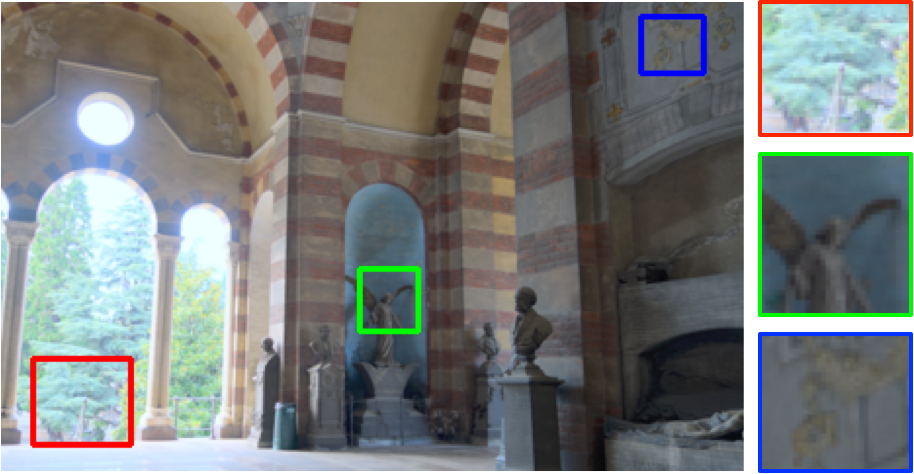}
    }
    \subfloat[LLF~\cite{paris2011local}]{
        \includegraphics[width=0.24\textwidth]{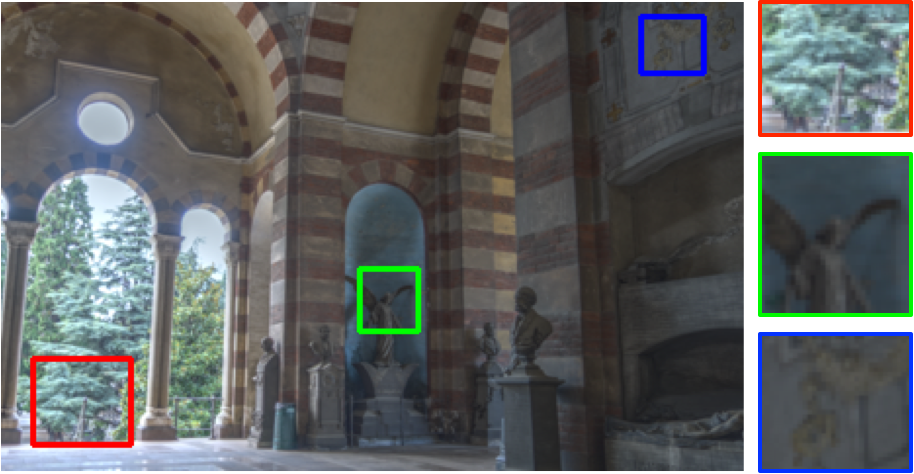}
    }
    \subfloat[GR~\cite{shibata2016gradient}]{
        \includegraphics[width=0.24\textwidth]{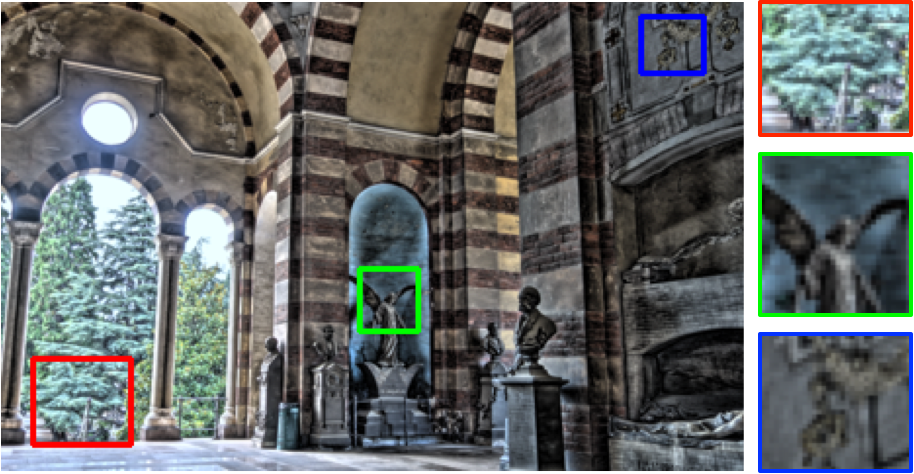}
    }
    \subfloat[Ours]{
        \includegraphics[width=0.24\textwidth]{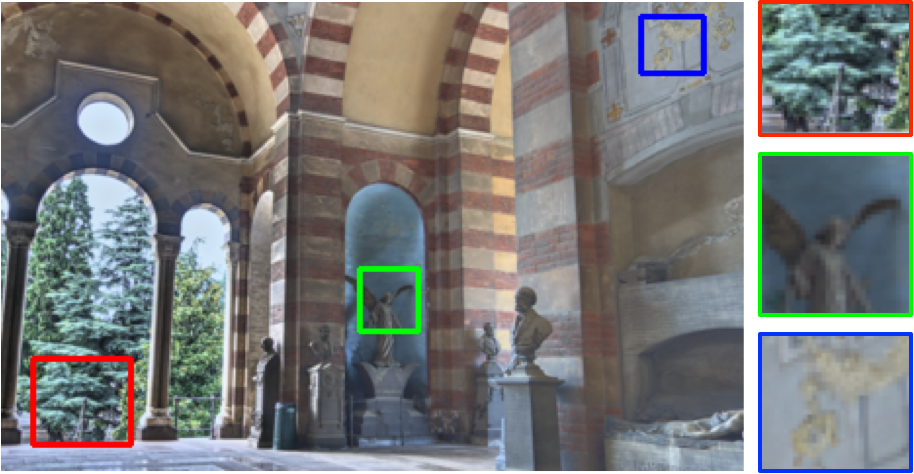}
    }
    \vspace{.2em}
    \centering
    \caption{Tone mapping results of the ``Architecture'' image courtesy of Nemoto Hiromi.}
	\label{fig:indoor}
\end{figure*}


\begin{figure*}[t]
    \centering
    \subfloat[Kim08~\cite{kim2008consistent}]{
        \includegraphics[width=0.24\textwidth]{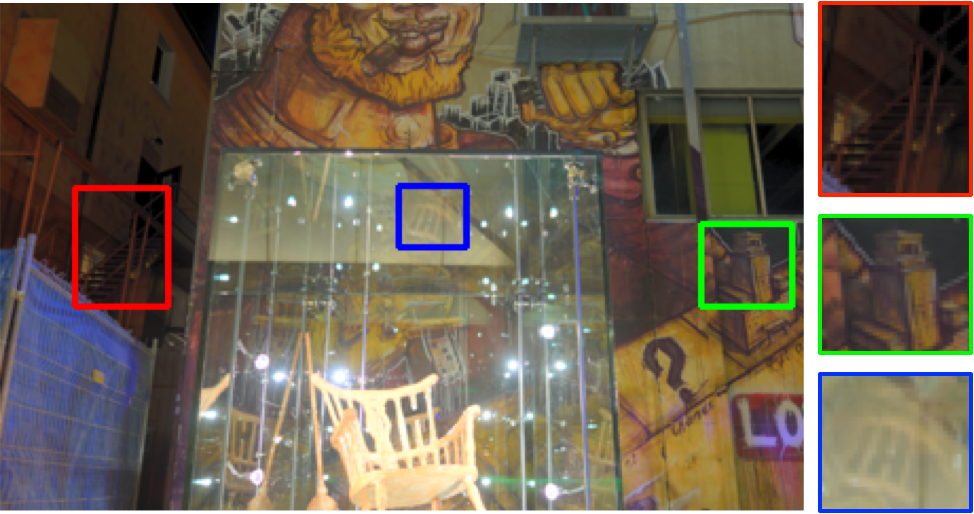}
    }
    \subfloat[Liang18~\cite{liang2018hybrid}]{
        \includegraphics[width=0.24\textwidth]{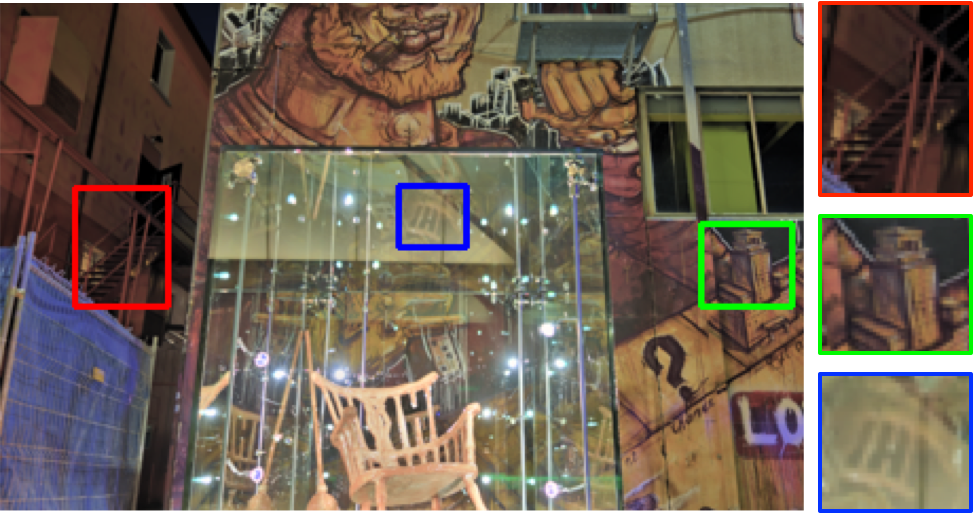}
    }
    \subfloat[NLPD-Opt~\cite{laparra2017perceptually}]{
        \includegraphics[width=0.24\textwidth]{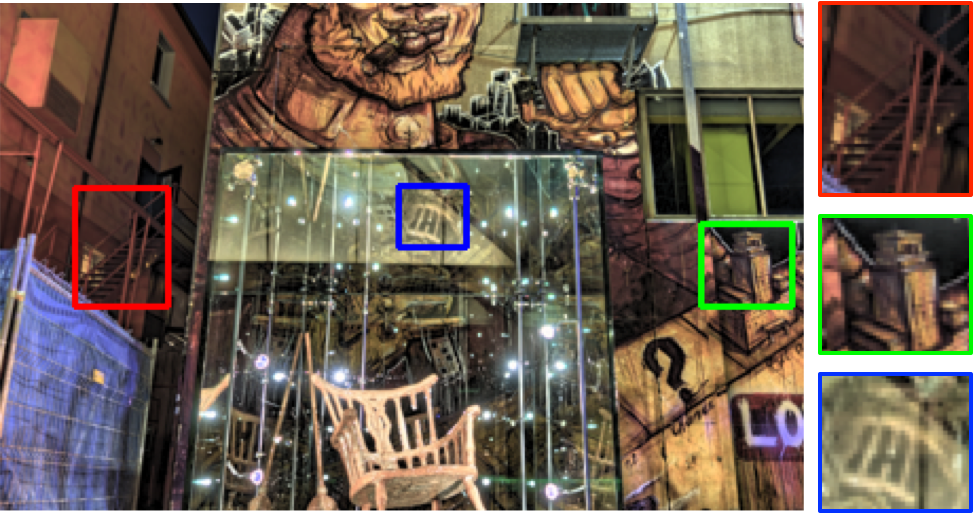}
    }
    \subfloat[Ours]{
        \includegraphics[width=0.24\textwidth]{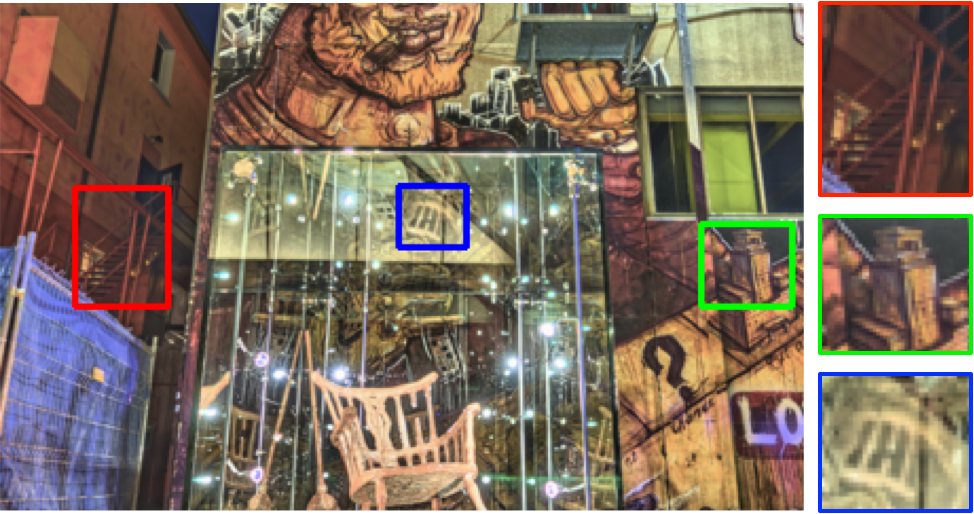}
    }
    \vspace{.2em}
    \caption{Tone mapping results of the ``Night Building'' image courtesy of Nemoto Hiromi.}
	    \label{fig:night}
\end{figure*}

A noticeable difference compared to the original CAN~\cite{chen2017fast} is that we remove all the bias terms, including those used in adaptive normalization. As shown in~\cite{mohan2019robust}, a bias-free neural network with LReLU non-linearly is \textit{locally} scale-invariant: rescaling the input by a constant value simply rescales the output by the same amount~\cite{mohan2019robust}
\begin{align}
    g(\alpha y_j^{(i)}) = \alpha g(y_j^{(i)}), \quad j\in\{1,\ldots, n^{(i)}\}.    
\end{align}
Scale-invariance renders CAN more robust to various luminance levels provided that natural scenes with different dynamic ranges are present in the training set. 
    
\subsection{Model Training and Testing}
\label{train_and_test}

We decompose the HDR image into a five-level normalized Laplacian pyramid. For the objective function NLPD~\cite{laparra2017perceptually}, we follow the original paper, and set the front-end non-linearity $\gamma$ to $1/2.6$, the local weight function $P$ to a spatially separable five-tap filter $[0.05, 0.25, 0.4, 0.25, 0.05]$ and the additive constant $c$ to  $0.17$ for bandpass channels, $P$ to $I$ and $c$ to $4.86$ for the lowpass channel, the two exponents for the metric $\alpha$ to $2.0$ and $\beta$ to $0.6$, respectively. The slope $\lambda_3$ in the LReLU is set to $0.2$.

During training, we use the Adam optimizer~\cite{Kingma2014adam} with a mini-batch size of $4$. The initial learning rate is set to $ 10^{-3}$ with a decay  factor of $10$ for every $1,000$ epochs, and we train our method for $2,000$ epochs. We 
calibrate HDR images by randomly sampling the maximum luminance values from $10^3$ to $10^5$ cd/m$^2$. In addition, we augment the training data by random cropping and horizontal flipping. During testing, we resize each HDR image such that the short side has a size of $512$, and make an empirical guess of the maximum luminance $S_{\mathrm{max}}$ in the original scene. 

\section{Experiments}
\label{sec:experiment}

In this section, we conduct experiments to demonstrate the promise of the proposed TMO. We first collect a database of $432$ HDR scenes, and use $391$ images for training and the rest  for testing. 

We select nine TMOs for comparison, including Drago03~\cite{drago2003adaptive}, Reinhard02~\cite{reinhard2002photographic}, Kim08~\cite{kim2008consistent}, WLS~\cite{farbman2008edge}, LLF~\cite{paris2011local}, Bruce14~\cite{BRUCE201412}, GR~\cite{shibata2016gradient}, NLPD-Opt~\cite{laparra2017perceptually}, and Liang18~\cite{liang2018hybrid}. Among these methods, Drago03, Reinhard02, and Kim08 are global operators, while WLS, LLF, Bruce14, GR, NLPD-Opt, and Liang18 are local operators. 
It is noteworthy that NLPD-Opt directly minimizes the NLPD metric in the image space. Therefore, given sufficient iterations, it can be regarded as the lower bound for all TMOs in terms of NLPD~\cite{laparra2017perceptually}. 
The implementations for all algorithms are obtained from the respective authors and  tested with the default settings.

\subsection{Qualitative Comparison} 

Fig.~\ref{fig:lamp} shows  the tone mapping results of the ``Lamp" HDR scene. The na\"{i}ve linear scaling creates a dark background with loss of details. The local contrast of the image by Drago03~\cite{drago2003adaptive} is significantly reduced (\eg, texts in the book). WLS~\cite{farbman2008edge} successfully preserves the structures in dark areas, but suffers from the over-exposure problem in bright regions. In contrast, our method produces a more natural appearance with rich details.

Fig.~\ref{fig:indoor} shows  the tone mapping results of the ``Architecture" HDR image. The bright regions of the image by Dargo03~\cite{reinhard2005dynamic} is a little over-exposed. GR~\cite{shibata2016gradient} tends to overshoot local details, making the image artificial. The proposed method generates a warmer appearance than that of LLF. Nevertheless, they produce close visual results with little artifact.


Fig.~\ref{fig:night} shows the tone mapping results of the ``Night Building'' HDR image. The image by Kim18~\cite{kim2008consistent} exhibits reduced global contrast due to the extreme dynamic range of the scene. Liang18~\cite{liang2018hybrid} succeeds in improving the details of the vitrine and the background, whose contrast is further improved by our method. With a substantially lower computational cost, our result closely matches that of NLPD-Opt~\cite{laparra2017perceptually}.

\begin{table}[t]
	\begin{center}
	\caption{Quantitative results of our method against existing TMOs in terms of TMQI~\cite{yeganeh2012objective} (and its two components structural fidelity $F$ and statistical naturalness $N$), NLPD~\cite{laparra2017perceptually}, and running time (in seconds) on test images with the short side resized to $512$. Local TMOs are highlighted in \textit{italics}.} \label{tab:tmqi_nlpd_time_score}
	   \vspace{1em}
		\begin{tabular}{l|c c c c c }
			\toprule
			TMO & TMQI$\uparrow$ & $F\uparrow$ & $N\uparrow$ & NLPD$\downarrow$ & Time\\
			\hline
			Drago03 & 0.887 & 0.903 & 0.467 & 0.240 & {\bfseries 0.21}\\
			Reinhard02& 0.875 & 0.888 & 0.420 & 0.230 & 0.24\\
			Kim08 & 0.893 & 0.912 & 0.488 & 0.219 & 0.62\\
			\textit{WLS} & 0.906 & 0.898 & 0.613 & 0.212 & 3.25\\
			\textit{LLF} & 0.913 & {\bfseries 0.918} & 0.622 & 0.197 & 356.81\\
			\textit{Bruce14} & 0.852 & 0.831 & 0.384 & 0.320 & 9.14\\
			\textit{GR} & 0.866 & 0.876 & 0.393 & 0.231 & 12.56\\
			\textit{NLPD-Opt} & 0.908 & 0.911 & 0.558 & {\bfseries 0.168} & 160.40\\
			\textit{Liang18} & 0.910 & 0.892 & 0.611 & 0.219 & 1.61\\
			\hline
			\textit{Ours} & {\bfseries 0.925} & 0.906 & {\bfseries 0.667} &  0.175 & 1.44 \\
		\bottomrule
		\end{tabular}
	\end{center}
\end{table}
 
\begin{figure*}[t]
\centering
\subfloat[One-level]{
\includegraphics[width=0.19\textwidth]{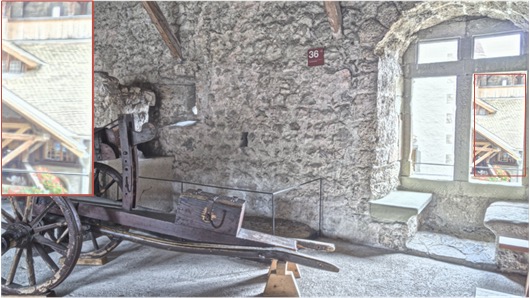}
}
\subfloat[Two-level]{
\includegraphics[width=0.19\textwidth]{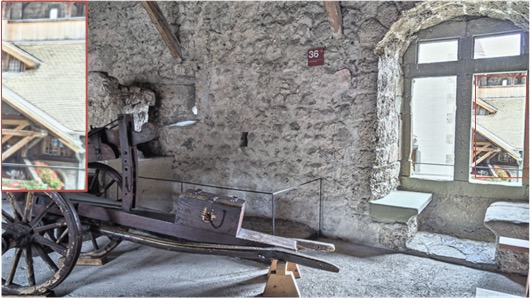}
}
\subfloat[Three-level]{
\includegraphics[width=0.19\textwidth]{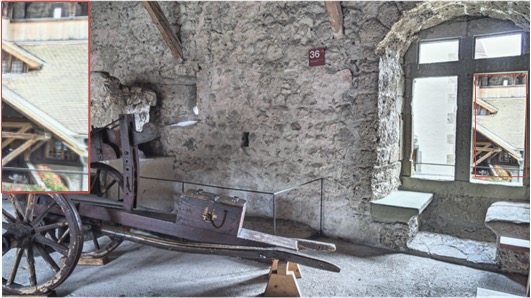}
}
\subfloat[Four-level]{
\includegraphics[width=0.19\textwidth]{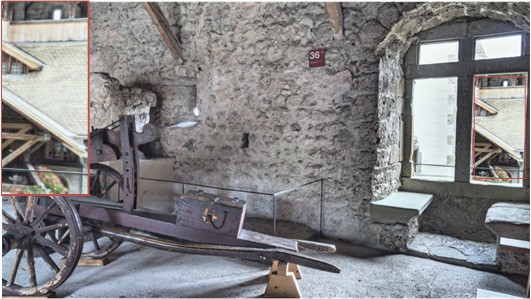}
}
\subfloat[Five-level]{
\includegraphics[width=0.19\textwidth]{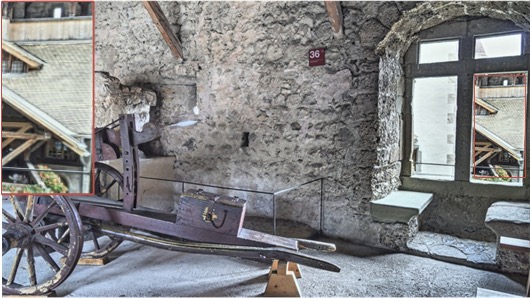}
}
\vspace{.2em}
\caption{Tone mapping results of the ``Workshop'' image with different input pyramid levels. Image courtesy of Nemoto Hiromi.}
\label{fig:multi_scale}
\end{figure*}

\begin{figure*}[t]
\centering
\subfloat[MAE-optimized]{
\includegraphics[width=0.24\textwidth]{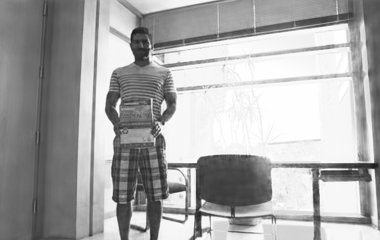}
}
\subfloat[SSIM-optimized~\cite{wang2004image}]{
\includegraphics[width=0.24\textwidth]{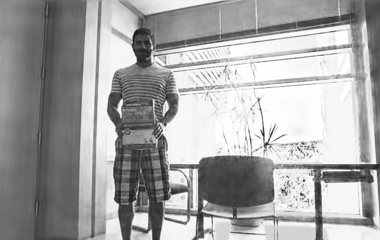}
}
\subfloat[TMQI-optimized~\cite{yeganeh2012objective}]{
\includegraphics[width=0.24\textwidth]{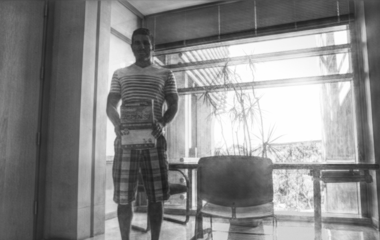}
}
\subfloat[NLPD-optimized~\cite{laparra2017perceptually}]{
\includegraphics[width=0.24\textwidth]{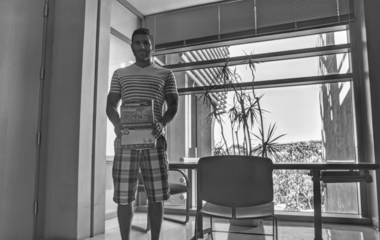}
}
\vspace{.2em}
\caption{Tone mapping results of the ``Man'' image with different objective functions. Image courtesy of Nima Khademi Kalantari.}
\label{fig:Opt-compare}
\end{figure*}
\subsection{Quantitative Comparison}

 We adopt two objective metrics for quantitative performance evaluation: TMQI~\cite{yeganeh2012objective} and NLPD~\cite{laparra2017perceptually}. TMQI is a variant of the SSIM index~\cite{wang2004image} for comparing images of different dynamic ranges. It combines structural fidelity (denoted by $F$) and statistical naturalness (denoted by $N$)  measurements to evaluate the visual quality of a tone-mapped image using the corresponding HDR image as reference. A larger TMQI or a smaller NLPD value indicates better perceptual quality. Table~\ref{tab:tmqi_nlpd_time_score} shows the results, from which we find that local operators generally outperform global operators in terms of TMQI. This is not surprising because TMQI is biased towards comparing local structure similarity, which is the design focus of local TMOs. This result is less obvious in terms of NLPD. As expected, NLPD-Opt achieves the best NLPD performance, followed by the proposed TMO and LLF. It is interesting to see that although our method is guided by NLPD, it achieves the best performance measured by TMQI. This provides strong justifications for our architectural design.
 
We test the running time of our method with existing TMOs on a computer with a 4.4GHz CPU and 64G RAM. The proposed TMO is implemented using PyTorch, while all competing methods are based on MATLAB\footnote{\url{https://github.com/banterle/HDR_Toolbox}} implemented by Banterle~\etal~\cite{Banterle:2017}. From Table \ref{tab:tmqi_nlpd_time_score}, we observe that our method is the fastest local TMO, which is  attributed to the manually optimized network architecture with only $74,378$ parameters. Furthermore, when an NVIDIA GTX 2080Ti GPU is enabled, our method runs the fastest among all methods ($0.017$ seconds).

\subsection{Ablation Analysis} 

We conduct ablation experiments to single out the contribution of the normalized Laplace decomposition and the perceptual optimization of the proposed TMO. We first analyze the effect of the input pyramid level on final visual quality. Note that one level corresponds to directly feeding the raw HDR image into a single network for tone mapping. As shown in Fig.~\ref{fig:multi_scale}, more levels lead to improved detail reproduction at the cost of  increased computational complexity. The default five-level pyramid keeps a good balance between visual quality and computational speed.   We then switch NLPD \cite{laparra2017perceptually} to three other objective functions: mean absolute error (MAE), SSIM \cite{wang2004image}, and TMQI \cite{yeganeh2012objective}, while fixing the network architecture. Fig.~\ref{fig:Opt-compare} shows the optimization results, which are optimal under their respective objectives. As can be seen, NLPD-optimized network achieves the best visual result.



\section{Conclusion}
\label{sec:conclusion}

We have introduced a perceptually optimized TMO based on light-weight DNNs. We find that the optimized method matches or exceeds the state-of-the-art across a variety of HDR natural scenes, which is verified by another perceptual quality metric, TMQI.

As with all biologically-inspired TMOs, our method requires specifying the maximum luminance value $S_\mathrm{max}$ during training and testing. Note that directly optimizing NLPD would lead to the trivial solution of $S_\mathrm{max}=0$, creating a completely black appearance. A statistical naturalness measure (\eg, implemented using a no-reference image quality model) may be added to help the optimization get rid of such bad local minima.

In our experiments, we assume a fixed display constraint with a minimum luminance of $I_{\mathrm{min}}=5$ cd/m$^2$ and  a maximum luminance of $I_{\mathrm{max}}=300$ cd/m$^2$. In the future, we will take steps to incorporate various display constraints into the proposed perceptual optimization framework.

\bibliographystyle{IEEEtran}
\bibliography{main}

\begin{thebibliography}{10}
\providecommand{\url}[1]{#1}
\csname url@samestyle\endcsname
\providecommand{\newblock}{\relax}
\providecommand{\bibinfo}[2]{#2}
\providecommand{\BIBentrySTDinterwordspacing}{\spaceskip=0pt\relax}
\providecommand{\BIBentryALTinterwordstretchfactor}{4}
\providecommand{\BIBentryALTinterwordspacing}{\spaceskip=\fontdimen2\font plus
\BIBentryALTinterwordstretchfactor\fontdimen3\font minus
  \fontdimen4\font\relax}
\providecommand{\BIBforeignlanguage}[2]{{%
\expandafter\ifx\csname l@#1\endcsname\relax
\typeout{** WARNING: IEEEtran.bst: No hyphenation pattern has been}%
\typeout{** loaded for the language `#1'. Using the pattern for}%
\typeout{** the default language instead.}%
\else
\language=\csname l@#1\endcsname
\fi
#2}}
\providecommand{\BIBdecl}{\relax}
\BIBdecl

\bibitem{reinhard2010high}
E.~Reinhard, W.~Heidrich, P.~Debevec, S.~Pattanaik, G.~Ward, and K.~Myszkowski,
  \emph{High Dynamic Range Imaging: Acquisition, Display, and Image-Based
  Lighting}.\hskip 1em plus 0.5em minus 0.4em\relax Morgan Kaufmann, 2010.

\bibitem{reinhard2002photographic}
E.~Reinhard, M.~Stark, P.~Shirley, and J.~Ferwerda, ``Photographic tone
  reproduction for digital images,'' \emph{ACM Transaction on Graphics},
  vol.~21, no.~3, pp. 267--276, 2002.

\bibitem{durand2002fast}
F.~Durand and J.~Dorsey, ``Fast bilateral filtering for the display of
  high-dynamic-range images,'' \emph{ACM SIGGRAPH Computer Graphics}, vol.~21,
  no.~3, pp. 257--266, 2002.

\bibitem{li2005compressing}
Y.~Li, L.~Sharan, and E.~H. Adelson, ``Compressing and companding high dynamic
  range images with subband architectures,'' \emph{ACM Transaction on
  Graphics}, vol.~24, no.~3, pp. 836--844, 2005.

\bibitem{meylan2006high}
L.~Meylan and S.~Susstrunk, ``High dynamic range image rendering with a
  retinex-based adaptive filter,'' \emph{IEEE Transactions on Image
  Processing}, vol.~15, no.~9, pp. 2820--2830, 2006.

\bibitem{farbman2008edge}
Z.~Farbman, R.~Fattal, D.~Lischinski, and R.~Szeliski, ``Edge-preserving
  decompositions for multi-scale tone and detail manipulation,'' \emph{ACM
  Transaction on Graphics}, vol.~27, no.~3, pp. 67:1--67:10, 2008.

\bibitem{ward1994radiance}
G.~J. Ward, ``The {RADIANCE} lighting simulation and rendering system,'' in
  \emph{Computer Graphics and Interactive Techniques}, 1994, pp. 459--472.

\bibitem{larson1997visibility}
G.~W. Larson, H.~Rushmeier, and C.~Piatko, ``A visibility matching tone
  reproduction operator for high dynamic range scenes,'' \emph{IEEE
  Transactions on Visualization and Computer Graphics}, vol.~3, no.~4, pp.
  291--306, 1997.

\bibitem{tumblin1993tone}
J.~Tumblin and H.~Rushmeier, ``Tone reproduction for realistic images,''
  \emph{IEEE Computer Graphics and Applications}, vol.~13, no.~6, pp. 42--48,
  1993.

\bibitem{drago2003adaptive}
F.~Drago, K.~Myszkowski, T.~Annen, and N.~Chiba, ``Adaptive logarithmic mapping
  for displaying high contrast scenes,'' \emph{Computer Graphics Forum},
  vol.~22, no.~3, pp. 419--426, 2003.

\bibitem{reinhard2005dynamic}
E.~Reinhard and K.~Devlin, ``Dynamic range reduction inspired by photoreceptor
  physiology,'' \emph{IEEE Transactions on Visualization and Computer
  Graphics}, vol.~11, no.~1, pp. 13--24, 2005.

\bibitem{kim2008consistent}
M.~H. Kim and J.~Kautz, ``Consistent tone reproduction,'' in
  \emph{International Conference on Computer Graphics and Imaging}, 2008, pp.
  152--159.

\bibitem{paris2011local}
S.~Paris, S.~W. Hasinoff, and J.~Kautz, ``Local {Laplacian} filters: Edge-aware
  image processing with a {Laplacian} pyramid.'' \emph{ACM Transaction on
  Graphics}, vol.~30, no.~4, pp. 68:1--68:12, 2011.

\bibitem{BRUCE201412}
N.~D. Bruce, ``{ExpoBlend}: Information preserving exposure blending based on
  normalized log-domain entropy,'' \emph{Computers \& Graphics}, vol.~39, pp.
  12--23, 2014.

\bibitem{shibata2016gradient}
T.~Shibata, M.~Tanaka, and M.~Okutomi, ``Gradient-domain image reconstruction
  framework with intensity-range and base-structure constraints,'' in
  \emph{IEEE Conference on Computer Vison and Pattern Recognition}, 2016, pp.
  2745--2753.

\bibitem{liang2018hybrid}
Z.~Liang, J.~Xu, D.~Zhang, Z.~Cao, and L.~Zhang, ``A hybrid $\ell_1$-$\ell_0$
  layer decomposition model for tone mapping,'' in \emph{IEEE Conference on
  Computer Vison and Pattern Recognition}, 2018, pp. 4758--4766.

\bibitem{yeganeh2013high}
H.~Yeganeh and Z.~Wang, ``High dynamic range image tone mapping by maximizing a
  structural fidelity measure,'' in \emph{IEEE International Conference on
  Acoustics, Speech, and Signal Processing}, 2013, pp. 1879--1883.

\bibitem{yeganeh2012objective}
------, ``Objective quality assessment of tone-mapped images,'' \emph{IEEE
  Transactions on Image Processing}, vol.~22, no.~2, pp. 657--667, 2012.

\bibitem{ma2015high}
K.~Ma, H.~Yeganeh, K.~Zeng, and Z.~Wang, ``High dynamic range image compression
  by optimizing tone mapped image quality index,'' \emph{IEEE Transactions on
  Image Processing}, vol.~24, no.~10, pp. 3086--3097, 2015.

\bibitem{laparra2017perceptually}
V.~Laparra, A.~Berardino, J.~Ball{\'e}, and E.~P. Simoncelli, ``Perceptually
  optimized image rendering,'' \emph{Journal of the Optical Society of America
  A}, vol.~34, no.~9, pp. 1511--1525, 2017.

\bibitem{burt1983laplacian}
P.~Burt and E.~Adelson, ``The {Laplacian} pyramid as a compact image code,''
  \emph{IEEE Transactions on Communications}, vol.~31, no.~4, pp. 532--540,
  1983.

\bibitem{yu2015multi}
F.~Yu and V.~Koltun, ``Multi-scale context aggregation by dilated
  convolutions,'' in \emph{International Conference on Learning
  Representations}, 2016, pp. 1--13.

\bibitem{chen2017fast}
Q.~Chen, J.~Xu, and V.~Koltun, ``Fast image processing with fully-convolutional
  networks,'' in \emph{IEEE International Conference on Computer Vision}, 2017,
  pp. 2497--2506.

\bibitem{mohan2019robust}
S.~Mohan, Z.~Kadkhodaie, E.~P. Simoncelli, and C.~Fernandez-Granda, ``Robust
  and interpretable blind image denoising via bias-free convolutional neural
  networks,'' in \emph{International Conference on Learning Representations},
  2020, pp. 1--10.

\bibitem{Kingma2014adam}
D.~P. Kingma and J.~Ba, ``Adam: {A} method for stochastic optimization,''
  \emph{CoRR}, vol. abs/1412.6980, 2014.

\bibitem{wang2004image}
Z.~Wang, A.~C. Bovik, H.~R. Sheikh, and E.~P. Simoncelli, ``Image quality
  assessment: From error visibility to structural similarity,'' \emph{IEEE
  Transactions on Image Processing}, vol.~13, no.~4, pp. 600--612, 2004.

\bibitem{Banterle:2017}
F.~Banterle, A.~Artusi, K.~Debattista, and A.~Chalmers, \emph{Advanced High
  Dynamic Range Imaging (2nd Edition)}.\hskip 1em plus 0.5em minus 0.4em\relax
  Natick, MA, USA: AK Peters (CRC Press), 2017.

\end{thebibliography}
\end{document}